\documentclass[runningheads]{llncs}
\usepackage{hyperref}
\usepackage{makeidx} 

\usepackage{amssymb, amsmath}
\usepackage[ruled,vlined]{algorithm2e}
\usepackage{graphicx}
\usepackage{multirow}

\usepackage{array}

\newcolumntype{C}[1]{>{\centering\let\newline\\\arraybackslash\hspace{0pt}}m{#1}}

\begin{document}
\frontmatter 
\pagestyle{headings} 
\addtocmark{Organizing Multimedia Data in Video Surveillance Systems Based on Face Verification with Convolutional Neural Networks} 

\mainmatter 
\title{Organizing Multimedia Data in Video Surveillance Systems Based on Face Verification with Convolutional Neural Networks}
\titlerunning {Organizing Multimedia Data in Video Surveillance Systems}
%
\author{Anastasiia D. Sokolova, Angelina S. Kharchevnikova, Andrey V. Savchenko}
\authorrunning{A.D.Sokolova, A.S.Kharchevnikova, A.V. Savchenko} 
%
\tocauthor{Anastasiia~Sokolova, Angelina~Kharchevnikova, Andrey~Savchenko}
\institute{National Research University Higher School of Economics, Nizhny Novgorod, Russian Federation\\
\email{adsokolova96@mail.ru}}


\maketitle 

\begin{abstract}
In this paper we propose the two-stage approach of organizing information in video surveillance systems. At first, the faces are detected in each frame and a video stream is split into sequences of frames with face region of one person. Secondly, these sequences (tracks) that contain identical faces are grouped using face verification algorithms and hierarchical agglomerative clustering. Gender and age are estimated for each cluster (person) in order to facilitate the usage of the organized video collection. The particular attention is focused on the aggregation of features extracted from each frame with the deep convolutional neural networks. The experimental results of the proposed approach using YTF and IJB-A datasets demonstrated that the most accurate and fast solution is achieved for matching of normalized average of feature vectors of all frames in a track.

\end{abstract}
\section{Introduction}

Nowadays, due to the growth of the multimedia data volume, the task of forming an automatic approach to the ordering of digital information is attracting increasing attention \cite{1}. The various photo organizing systems allows the user to speed up the search for the required frame, and also to increase the efficiency of work with the media library. Such modern solutions include services like Apple iPhoto, Google Photos, etc., which are designed to store, organize and display media data. However, multimedia data organization systems are required not only for a particular user who has an archive of photographs, but also for the field of public safety, where video surveillance technologies are used for monitoring purposes \cite{2}. Consequently, there is a challenge of ordering the visitors, whose faces are observed in a surveillance system. To solve the problem the clustering of video tracks that contains the same person can be performed using the known face verification methods \cite{3,4} based on deep convolutional neural networks (CNNs) \cite{5,6,7,8}. Unlike the traditional technologies of ordering digital information, video surveillance systems are characterized by a large amount of data, because hundred frames can be obtained in dynamics in a few seconds \cite{4,9}. Therefore the goal of our research is to improve the verification efficiency by the combination for features extracted from individual frames. 
The rest of the paper is organized as follows: in Section 2, we formulate the pro-posed approach of organizing multimedia data in video surveillance system. In Section 3, we present the experimental results in unconstrained face verification. Concluding comments are given in Section 4.

\section{Automatic Organization of Video Data}

The task of this paper is to split the given video sequence of T frames into subsequences with observations of one person, and then unite different subsequences containing the same person. At first, the facial regions are detected \cite{9} using, e.g., the Viola-Jones method. For simplicity, we assume that each frame in the given video consists of images (frames) of exactly one face. Next, an appropriate tracker algorithm  \cite{9,10} divides the input sequence into $M<T$ disjoint subsequences (tracks) $\{X(m)\}$, $m = 1,2, ..., M$, where the $m$-th frame is characterized by its borders ($t_1(m)$,$t_2(m)$), where the $m$-th track contains $\mathrm{\Delta t(m)=t_2(m)-t_1(m)+1}$ frames. Finally, we search for similar tracks using, e.g., hierarchical agglomerative clustering methods \cite{11}: similar objects are sequentially grouped together. In order to implement any clustering method, a dissimilarity measure between video tracks should be defined. Let us extract appropriate facial features from every frame.  

Nowadays feature extraction is implemented using the deep CNNs trained with an external large dataset, e.g., Casia WebFaces or MS-Celeb-1M \cite{6,12}. The outputs of the CNN’s last (bottleneck) layer for the $t$-th frame are stored in the D-dimensional feature vector $x(t)$. These bottleneck features are usually matched with the Euclidean distance $ρ(x(t_1), x(t_2))$ \cite{12}. It is possible to define the dissimilarity of tracks $X(m_1)$ and $X(m_2)$ as a summary statistic of distances between individual frames. In our experiments the highest accuracy was achieved with the average distance:

\begin{equation}
\rho(\textbf{$X$}(m_1), \textbf{$X$}(m_2))=\frac{1}{\Delta t(m_1)\Delta t(m_2)}\sum\limits_{t=t_1(m_1)}^{t_2(m_1)}\sum\limits_{t'=t_1(m_2)}^{t_2(m_2)}\rho(x(t),x(t')).
\end{equation}

However, the run-time complexity of such distance is high due to the pair-wise matching of all frames in these tracks causing the computation of $\mathrm{\Delta t(m_1)\Delta t(m_2)}$ distances between high-dimensional features. Hence, in this paper we examine the computation of the distance between tracks $X(m_1)$ and $X(m_2)$ as the distance between their fixed-size representations. Yang et al  \cite{5} proposed the 2-layer neural network with attention blocks to aggregate the CNN features of all frames. However, in our experiments the robustness of this approach was insufficient, hence, we use straight-forward aggregation (or pooling \cite{5}) techniques:

\begin{enumerate}

     \item The distance between tracks is defined as the distance between their medoids:
    \begin{multline}
     \rho(\textbf{$X$}(m_1), \textbf{$X$}(m_2))=\rho(\textbf{x*}(m_1), \textbf{x*}(m_2)), \\
    \textbf{x*}(m_i)=  \underset{x(t),t \in [t_1(m_i),t_2(m_i)]}{\operatorname{argmin}}\sum\limits_{t'=t_1(m_i)}^{t_2(m_i)}\rho(x(t),x(t')), i \in \{1,2\}.
   \end{multline}

    \item Average features of each track are matched:
    \begin{equation}
    \rho(\textbf{X}(m_1), \textbf{X}(m_2))=\rho(\bar{x}(m_1), \bar{x}(m_2)),\bar{x}(m_i)=\frac{1}{\Delta t(m_i)}\sum\limits_{t'=t_1(m_i)}^{t_2(m_i)}\textbf{x}(t). 
    \end{equation}
    
\end{enumerate}

It is worth noting that in static image recognition tasks the CNN bottleneck feature vectors are typically divided into their $\mathrm{L_2}$ norm \cite{12}. Such normalization is known to make these features more robust to variations of observation conditions, e.g., camera resolution, illumination and occlusion. However, in our task the sequences of frames are matched, so it is possible to slightly defer the normalization. Thus, in this paper we consider either conventional approach with aggregation of the normalized features (hereinafter ``$\mathrm{L_2}$-norm  -$>$ Medoid" (2) and ``$\mathrm{L_2}$-norm -$>$ AvePool" (3)), or its slightly modified version with normalization of aggregated vectors (2), (3) (hereinafter ``Medoid -$>$ $\mathrm{L_2}$-norm" and ``AvePool -$> \mathrm{L_2}$-norm", respectively).

We implemented the described approach in a special in MS Visual Studio 2015 project (github link will be provided after double-blind peer review) using C++ language and the OpenCV library, especially, its DNN and Tracking extra modules. The complete data flow in this system is presented in Fig. 1.

\begin{figure}[h]
 \includegraphics[width=1\textwidth]{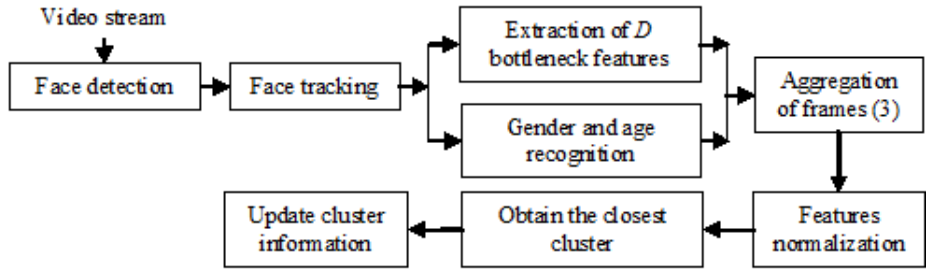}
 \caption{The data flow in the organizing video data system.}
 \label{fig1}
\end{figure}

Here we detect faces using the Viola-Jones cascades with the Haar features. The obtained facial regions are verified by additional eye detection \cite{9} and tracked using the KCF algorithm \cite{10}. Face detection is repeated periodically in order to: 1) verify the tracking results; 2) look for new faces, and 3) mark disappeared persons. In the latter case we extract D CNN bottleneck features for each frame of the track. These features are extended with the probabilities at the output of the gender and age prediction CNNs \cite{13}. In the data flow (Fig. 1) the simple online clustering of the normalized features is presented \cite{14}: the feature vector of the last track is matched with the features of previously detected clusters. If the distance to the nearest cluster does not exceed a certain threshold, this track is added to the cluster, and the information about the latter is updated. Namely, we compute the aggregated features (2) or (3) of a whole cluster and re-estimate the gender and age of persons discovered in the input video in order to facilitate the navigation through the organized collection of tracks. 

\section{Experimental Results }

In this section we provide experimental study of the key part of the proposed system (Fig. 1), namely, the matching of video tracks in unconstrained face verification task. In addition to described aggregation techniques (``$\mathrm{L_2}$-norm -$>$ Medoid" and ``Medoid -$>$ $\mathrm{L_2}$-norm" (2), ``$\mathrm{L_2}$-norm -$>$ AvePool" (3)) we examined the pairwise comparison of all frames in both tracks (1). Moreover, we implemented the described techniques (1)-(3) unnormalized features. To extract features, we used the Caffe framework and two publicly available CNNs suitable for face recognition, namely, the VGGNet \cite{6} and Lightened CNN (version C) \cite{12}. The VGGNet extracts $D$ = 4096 non-negative features in the output of ``fc7" layer from 224x224 RGB images. $D$ = 256 features (``eltwise\_fc2" layer) are computed when 128x128 grayscale image of the facial region is fed into the Lightened CNN. All experiments were performed on the computer Lenovo ideapad 310, 64-bit operating system with NVIDIA GeForce 920MX.

Our first experiments were conducted on the YouTube Faces (YTF) database \cite{15}, which contains 3,425 videos of 1,595 different people. An average of 2.15 videos are available for each subject. The shortest track duration is 48 frames, the longest track contains 6,070 frames, and the average length of a video clip is 181.3. The estimates of AUC (Area under curve) and FRR (False Reject Rate) for fixed FAR (False Accept Rate) using the YTF face verification protocol are presented in Table 1 (in the format mean ± standard deviation). Here we do not display a row for ``$\mathrm{L_2}$-norm -$>$ Medoid", because its results are identical to the ``Medoid -$>$ $\mathrm{L_2}$-norm" due to the independence of the computed medoid (2) on the order of normalization.

\begin{table}[h]
\caption{Results of video-based face verification, YTF dataset }
\begin{center}
\begin{tabular}{p{0.17\linewidth}p{0.12\linewidth}p{0.12\linewidth}p{0.15\linewidth}p{0.12\linewidth}p{0.12\linewidth}p{0.12\linewidth}}

\hline\noalign{\smallskip}
 & \multicolumn{3}{c}{Lightened CNN} & \multicolumn{3}{c}{VGGNet}  \\
\hline\noalign{\smallskip}
 & AUC(\%) & ERR(\%) & FRR@FAR = 1\% & AUC(\%) & ERR(\%) & FRR@FAR = 1\% \\
\hline\noalign{\smallskip}
{Distance (1)} &	90.7$\pm$0.6 & 15.7$\pm$0.2 & 77.0$\pm$8.4 & 83.3$\pm$0.8 &	24.0$\pm$0.3 & 85.8$\pm$9.0 \\
{$\mathrm{L_2}$-norm -$>$ Distance (1)} &	98.2$\pm$0.4 & 6.0$\pm$0.1 & 14.1$\pm$3.6 & 97.9$\pm$0.6 & 6.0$\pm$0.1 & 23.2$\pm$6.3 \\
{Medoid (2)} & 84.7$\pm$0.7 &	25.0$\pm$0.3 & 72.9$\pm$7.8 & 80.8$\pm$1.2 & 27.0$\pm$0.4 & 83.9$\pm$7.7 \\
{Medoid (2) -$>$ $\mathrm{L_2}$-norm} & 88.8$\pm$0.6 & 19.0$\pm$0.2 & 54.1$\pm$5.9 & 85.2$\pm$0.7 & 23.0$\pm$0.2 & 69.9$\pm$7.9 \\
{AvePool (3)} & 91.8$\pm$1.4 & 13.0$\pm$0.1 & 72.3$\pm$11.5 & 87.4$\pm$1.2 & 39.0$\pm$0.3 & 81.2$\pm$5.8 \\
{$\mathrm{L_2}$-norm -$>$ AvePool (3)} & 96.8$\pm$0.5 & 12.0$\pm$0.1 & 37.2$\pm$7.6 & 96.3$\pm$0.7 & 39.0$\pm$0.3 & 76.9$\pm$6.8 \\
{AvePool (3) -$>$ $\mathrm{L_2}$-norm} & 97.6$\pm$0.5 & 7.5$\pm$0.1 & 12.5$\pm$3.1 & 97.7$\pm$0.6 & 13.0$\pm$0.1 & 25.3$\pm$7.8 \\
\hline\noalign{\smallskip}
\end{tabular}
\end{center}
\end{table}

These results emphasize the need for proper normalization of feature vectors. The most efficient algorithm is to normalize the features of all frames and then find the average distance (1). The obtained state-of-the-art result is 0.982, the difference be-tween it and 0.988 \cite{5} is not statistically significant. However, the normalization of AvePool (3) features is characterized by practically the same quality, though it is much faster. The AUC for matching of medoids (2) is 10-12\% less when compared to the AUC of the AvePool (3). It is worth noting that the latter method is also 1-2\% more accurate than the conventional approach \cite{4,5} of averaging the preliminarily normalized features. 

In the next experiment we obtained a cluster threshold by fixing FAR = 1\% and using training set, then applied the clustering of all tracks from the YTF. By using the Lightened CNN features, 1800 clusters were identified, 20 of them contain videos of different persons. The application of the VGGNet feature extraction increases the number of clusters to 2000 with 30 incorrect clusters.

The first experiment was repeated for rough grouping of tracks with the persons of approximately identical age and same gender using the probabilities at the outputs of the pre-trained CNNs \cite{13}. Table 2 contains AUC achieved for matching of $D$ = 8 posterior probabilities of age categories, $D$ = 2 ((male/female) posterior probabilities) and the union of these two feature sets. We used $\mathrm{L_1}$-norm to treat the features as posterior probabilities and compared them with either Euclidean ($\mathrm{L_2}$) distance of the Kullback-Leibler (KL) divergence, which is assumed to be more suitable for comparison of discrete probability distributions. Here the prior feature normalization is not needed, as the outputs of the CNNs softmax layers are $\mathrm{L_1}$ normed. These results are much worse when compared to facial features from the previous experiment. Nevertheless, the age and gender features can be potentially used to refine the results obtained by conventional face verification techniques (Table 1) AUC is 8-19\% higher than the random guess.

\begin{table}[h]
\caption{AUC (\%) of video-based face verification, YTF dataset, age and gender features}
\label{tab2}
\begin{center}
\begin{tabular}{ccccc}

\hline\noalign{\smallskip}
 & Distance & Age & Gender & Age and Gender \\
\hline\noalign{\smallskip}
\multirow{2}{*}{Distance (1)} & $\mathrm{L_2}$ & 60.8$\pm$1.3 & 65.8$\pm$0.8 & 68.7$\pm$0.8 \\
 & KL & 61.6$\pm$1.1 & 65.8$\pm$1.0 & 65.8$\pm$0.9 \\
\multirow{2}{*}{Medoid (2)} & $\mathrm{L_2}$ &  58.4$\pm$1.4 & 63.3$\pm$1.0 & 64.9$\pm$1.0 \\
 & KL &  58.9$\pm$1.4 & 63.4$\pm$1.0 & 64.8$\pm$0.9 \\
\multirow{2}{*}{AvePool (3) -$>$ $\mathrm{L_2}$-norm} & $\mathrm{L_2}$ &  60.4$\pm$1.3 & 65.7$\pm$0.9 & 67.9$\pm$0.9 \\
 & KL &  63.2$\pm$1.2 & 65.3$\pm$0.9 & 68.8$\pm$0.8 \\
\hline\noalign{\smallskip}
\end{tabular}
\end{center}
\end{table}

The last experiment was conducted on the IARPA Janus Benchmark A (IJB-A) (IJB-A) dataset \cite{16} with 2043 videos of 500 identities. Table 3 contains the results of several best aggregation techniques in the face verification with bottleneck features extracted by VGGNet and Lightened CNN. 

\begin{table}[h]
\caption{Results of video-based face verification, IJB-A dataset}
\label{tab3}
\begin{center}
\begin{tabular}{p{0.17\linewidth}p{0.12\linewidth}p{0.12\linewidth}p{0.15\linewidth}p{0.12\linewidth}p{0.12\linewidth}p{0.12\linewidth}}

\hline\noalign{\smallskip}
 & \multicolumn{3}{c}{Lightened CNN} & \multicolumn{3}{c}{VGGNet}    \\
\hline\noalign{\smallskip}
& AUC(\%) & ERR(\%) & FRR@FAR = 1\% & AUC(\%) & ERR(\%) & FRR@FAR = 1\% \\
\hline\noalign{\smallskip}
$\mathrm{L_2}$-norm -$>$ Distance (1) &	87.9$\pm$0.5 & 20.5$\pm$0.9 & 67.9$\pm$3.3 & 97.5$\pm$0.4 & 8.0$\pm$0.4 & 30.3$\pm$4.6 \\
Medoid (2) -$>$ $\mathrm{L_2}$-norm & 76.6$\pm$0.4 & 30.0$\pm$1.2 & 77.1$\pm$4.0 & 92.4$\pm$0.7 & 15.5$\pm$0.6 & 50.0$\pm$6.9 \\
$\mathrm{L_2}$-norm -$>$ AvePool (3) & 79.6$\pm$0.7 & 27.8$\pm$0.8 & 67.5$\pm$4.4 & 96.1$\pm$0.4 & 13.6$\pm$0.8 & 40.0$\pm$4.4 \\
AvePool (3) -$>$ $\mathrm{L_2}$-norm & 88.2$\pm$0.4 & 20.0$\pm$0.4 & 59.3$\pm$2.9 & 97.7$\pm$0.3 & 8.0$\pm$0.3 & 26.8$\pm$4.2 \\
\hline\noalign{\smallskip}
\end{tabular}
\end{center}
\end{table}

In contrast to the first experiment, here the VGGNet \cite{6} is much more accurate than the Lightened CNN \cite{12}. Our conclusions about relative efficiency of discussed aggregation techniques remain similar to the previous experiments. However, this dataset highlights the superiority of the normalized average features (AvePool -$>$ $\mathrm{L_2}$-norm): it drastically improves AUC and FRR, when compared to traditional imple-mentation of average pooling in aggregation of video features \cite{3,4}. 

\section{Conclusion}

In this paper we considered the automatic organizing the data in video surveillance systems (Fig. 1). We particularly focused on the ways to efficiently compute the dissimilarity of video tracks by using rather simple aggregation techniques. We experimentally supported the claim that the most accurate and computationally cheap technique involves the $\mathrm{L_2}$-normed average vector of unnormalized frame features. It was noticed that the sequence of this two operations is very important. In fact, much more widely used aggregation of normalized features \cite{2} is usually less accurate (Table 3).

The main direction for further research is applying our approach in organizing data from real video surveillance systems. It is also important to examine more sophisticated distances between video tracks, e.g., metric learning \cite{17} or statistical homogeneity testing \cite{11}. If the number of observed persons is high, it is necessary to deal with insufficient performance of our simple online clustering by using, e.g., approximate nearest neighbor search \cite{14,7,8,18}. Moreover, we are planning to introduce the weighing for different features including age and gender probabilities to make our algorithm more accurate. In fact, the accuracy of the age and gender prediction CNNs \cite{13} is rather low, hence, it is necessary to implement contemporary CNN architectures including Inception or ResNets.

\textbf{Acknowledgements.}
The article was prepared within the framework of the Academic Fund Program at the National Research University Higher School of Economics (HSE) in 2017 (grant №17-05-0007) and by the Russian Academic Excellence Project ``5-100". Andrey V. Savchenko is partially supported by Russian Federation President grant no. MD-306.2017.9.

\end{document}